# Improving the K-means Algorithm using Improved Downhill Simplex Search

Ehsan Saboori
K.N Toosi University of Technology
Tehran, Iran
ehsansaboori@sina.kntu.ac.ir

Shafigh Parsazad, Anoosheh Sadeghi
Ferdowsi University, University of East London
Mashhad, Iran
London, UK
Shafigh.Parsazad@stu-mail.um.ac.ir

*Abstract*— the k-means algorithm is one of the well-known and most popular clustering algorithms. K-means seeks an optimal partition of the data by minimizing the sum of squared error with an iterative optimization procedure, which belongs to the category of hill climbing algorithms. As we know hill climbing searches are famous for converging to local optimums. Since k-means can converge to a local optimum, different initial points generally lead to different convergence cancroids, which makes it important to start with a reasonable initial partition in order to achieve high quality clustering solutions. However, in theory, there exist no efficient and universal methods for determining such initial partitions. In this paper we tried to find an optimum initial partitioning for k-means algorithm. To achieve this goal we proposed a new improved version of downhill simplex search, and then we used it in order to find an optimal result for clustering approach and then compare this algorithm with Genetic Algorithm base (GA), Genetic K-Means (GKM), Improved Genetic K-Means (IGKM) and k-means algorithms.

## I. INTRODUCTION

The k-means algorithm is one of the most popular clustering algorithms. K-means seeks an optimal partition of the data with an iterative procedure. The basic clustering procedure of k-means is summarized as follows:

- Initializing a K-partition randomly or based on some prior knowledge. Calculating the cluster prototype matrix
- Assigning each object in the data set to the nearest cluster
- Recalculating the cluster prototype matrix based on the current partition
- Repeating steps 2 and 3 until there is no change for each cluster.

The k-means is regarded as a staple of clustering methods due to its ease of implementation. It works well for many practical problems, particularly when the resulting clusters are compact and hyper-spherical in shape. The time complexity of k-means is $O_{(NKdT)}$, where $T$ is the number of iterations, $N$ number of objects, $K$ number of clusters. Since $K, d,$ and $T$ are usually much less than $N$, the time complexity of k-means is approximately linear. Therefore, k-means is a good selection for clustering large scale data sets.

While k-means has these desirable properties, it also suffers several major drawbacks, particularly the inherent limitations when hill climbing methods are used for optimization. These disadvantages of k-means attract a great deal of effort from different communities, and as a result, many variants of k-means have appeared to address these obstacles. To overcome this problem we used an improved downhill simplex search to find better optimum solutions. In sections II and III we will introduce random search and downhill simplex searches. We also discuss about the improved version of downhill simplex search in section III. In section IV we introduce the proposed algorithm. The comparisons are in part V. Finally we can see the conclusion in part VI. Fig 1 shows a sample of k-means clustering result. In this sample data are divided into 5 clusters.

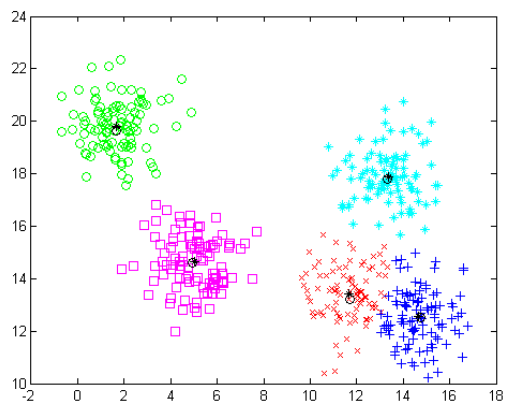

Figure 1 Sample K-means result for 5 clusters

## II. RANDOM SEARCH AND DOWN-HILL SIMPLEX SEARCH

### A. Random Search

Random search explores the parameter space of an objective function sequentially in a seemingly random fashion to find the optimal point that minimizes (or maximizes) the objective function. Besides being derivative free, the most distinguishing strength of the random search method lies in its simplicity, which makes the method easily understood and conveniently customized for specific applications. Random search steps:

- Choose a start point $x$ as the current point. Set initial bias b equal to a zero vector.
- Add a bias term $b$ and a random vector $dx$ to the current point $x$ in the input space and evaluate the objective function at the new point.

- If $f(x + b + dx) < f(x)$, set the current point *x* equal to $(x + b + dx)$ and the bias b equal to $0.2b + 0.4dx$
  Go to step 6. Otherwise, go to the next step.
- If $f(x + b - dx) < f(x)$, set the current point *x* equal to $(x + b - dx)$ and the bias b equal to $b - 0.4dx$
  Go to step 6. Otherwise, go to the next step.
- Set the bias equal to *0.5b* and go to step 6.
- Stop if the maximum number of function evaluations is reached. Otherwise go back to step 2 to find a new point.

Fig 2 shows the flowchart for random search.

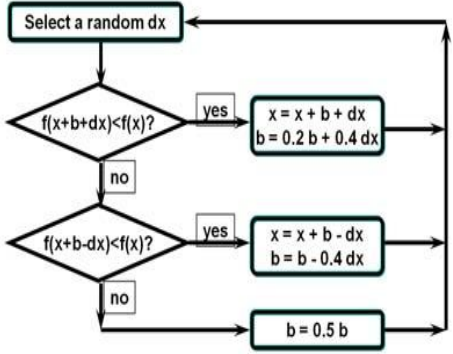

Figure 2. Random Search flowchart

### B. Down Hill Simplex search

Downhill simplex search is a derivative free method for multidimensional function optimization. In contrast to other derivative free approaches, this search method is not very efficient compared to derivative based methods. However, the concept behind downhill simplex search is simple and it has an interesting geometrical interpretation. We consider the minimization of a function of *n* variables with no constraints. We start with an initial simplex, which is a collection of *n + 1* point in *n* dimensional space. The downhill simplex search repeatedly replaces the point having the highest function value in a simplex with another point. Note that this method has little to do with the simple method for linear programming, except that both of them make use of the geometrical concept of a simplex. When combined with other operations, the simplex under consideration adapts itself to the local landscape, elongating down long inclined planes, changing direction on encountering a valley at an angle, and contracting in the neighborhood of a minimum [1]. To start the downhill simplex search, we must initialize a simplex of *n + 1* point. For example, a simplex is a triangle in two dimensional spaces and a tetrahedron in three dimensional spaces. Moreover, we would like the simplex to be no degenerate that is; it encloses a finite inner n-dimensional volume. Fig 3 shows the complete flowchart for downhill simplex search.

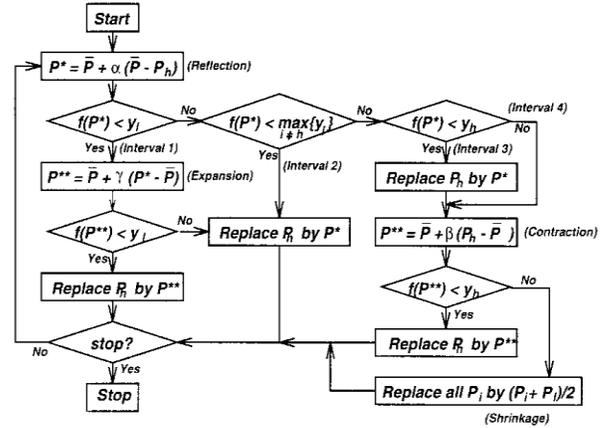

Figure 3. The flowchart of downhill simplex search

### III. IMPROVED DOWN HILL SIMPLEX SEARCH

Here we suggest a hint to improve this algorithm. As we know in this algorithm there are not any randomness and for same initial conditions we have same results and also when points get near algorithm approximately stops. We can solve these two problems by generating n random number when points get near each other and then continue with new n + 1 point. The fig 4 illustrates the improved downhill simplex search paradigm. The left image shows the simple downhill search and the right one shows the improved downhill search. As we can see, by using randomization, the algorithm can seek the optimal minimum or maximum.

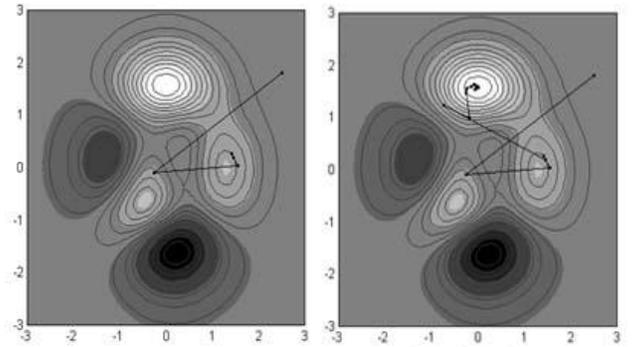

Figure 4. The improved downhill simplex search paradigm

Fig 5 shows the 3 dimensions environment for improved downhill simplex search. In this figure we can see the local minimum or maximum r peaks and simple downhill search maybe stick in local results. However the improved downhill simplex search can seeks the global minimum or maximum by jumping randomly in local result.

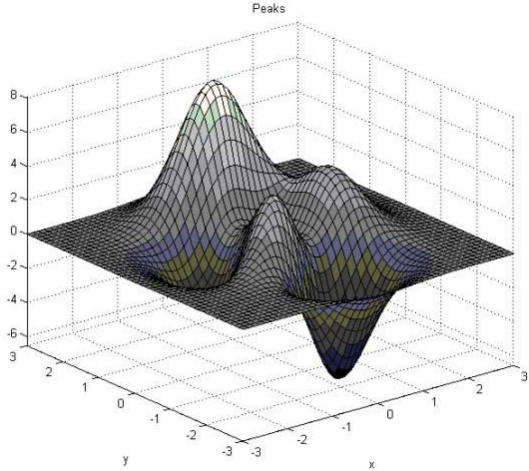

Figure 5. The 3 dimension environment for improved downhill simplex search

IV. THE PROPOSED ALGORITHM

As it mentioned before, the k-means algorithms do not seek optimal result in all the time. The result of this algorithm depends on the first means that was chosen randomly at first. So we can select the first means in such a manner that the result of k-means comes nearer to optimal result. In the proposed algorithm we use Improved Downhill Simplex Search to select the first $K$ means. Dunn index and Jagota index are used as cluster validity measures to compare efficiency of the proposed algorithm with other algorithms. Also the precision of accuracy is used for comparing. The Dunn index defines the ratio between the minimal intracluster distance to maximal intercluster distance. So the bigger Dunn index shows the better clustering we have.

$$Dunn_{index} = \frac{d_{min}}{d_{min}}$$

Jagota index defines the sum of the average distance in each cluster. So the better clustering has a smaller Jagota index.

$$Jagota_{index} = \sum_{k=0}^{n} d_k$$

Fig 6 illustrates the flowchart of the proposed algorithm. Improved Downhill Simplex Search is used to find suitable first means and then performs the k-means to cluster the dataset.

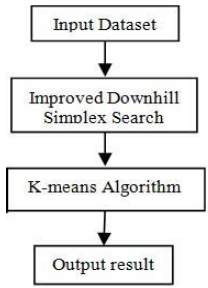

Figure 6. The proposed algorithm flowchart

V. THE COMPARISONS

To evaluate efficiency of proposed algorithm, we compare its accuracy precision of result with GA-based [2], Genetic K-means (GKM) [3], Improve Genetic K-means (IGKM) [4] and k-means algorithms. Also the Dunn and Jagota indexes are compared with k-means. For fairness comparison we use three real life data sets that are iris data sets ( $k_{optimal} = 3, n = 150$), glass data sets ($k_{optimal} = 6, n = 214$) and wine data set ($k_{optimal} = 3, n = 178$). We compare algorithms in two aspects, the first one is time complexity comparison and the other is the comparison of the effectiveness of the algorithms.

A. Time complexity

As the time complexity of IGKM is less than GA and GKM algorithms [5], we only compare the proposed algorithm, time complexity and IGKM. The time complexity of the proposed algorithm is much less than IGKM methods because in $n$ dimensional space we only have to compute evaluation function $n + 1$ times but in GA we should compute fitness for every chromosome. In IGKM the time complexity can be computed as:

$$IGKM_{Time\ Complexity} = p \times l \times k \times n \times s \times average\ (k\text{-}mean)$$

Where $p$ is number of population, $l$ is the number of GA's loops, $k$ is number of clusters, $n$ is number of data, $s$ is time of cross and mutation and *average (k-means)* is average of loops that k-means algorithm stops. The proposed approach's time complexity is:

$$Proposed\ Method_{Time\ Complexity} = c \times l \times k \times n \times d$$

Where $l$ is number of simplex search's loops, $c$ is number of attribute + 1, $n$ is number of data and $d$ is the depth of k-means algorithm. It shows how better the time complexity of the proposed algorithm is comparing to IGKM, because $d$ and $c$ are small variables.

B. Efficiency

To compare the efficiency of these algorithms, the precision of accuracy has been measured for k-means, GA, GKM, IGKM and the proposed algorithm. Table 1 illustrates the result of comparison for three different datasets. The proposed algorithm is less accurate than GKM and IGKM but due to its simplicity and speed it has competitive advantages to GKM and IGKM. The genetic algorithm consumes a lot of time to be performed.

TABLE 1. The comparison of the precision of accuracy

| Data sets | k-means | GA | GKM | IGKM | Proposed algorithm |
|---|---|---|---|---|---|
| Iris | 92.67% | 40% | 92.67% | 92.67% | 90% |
| Wine | 68.54% | 43.82% | 98.31% | 98.31% | 97.66% |
| Glass | 63.08% | 23.36% | 81.31% | 81.31% | 80.92% |

C. Compare with k-means

To compare the proposed method with k-means, we compare the Dunn and Jagota indexes. For iris and wine dataset the Dunn and Jagota indexes were calculated for both algorithms. Table 2 contains the result of Dunn index for k-means algorithm and the proposed algorithm. The Dunn index for the proposed algorithm is bigger than the k-means Dunn index; therefore it shows that the proposed algorithm improves the k-means clustering.

TABLE 2. The result of Dunn index for k-means and the proposed algorithm

| Data sets | k-means | proposed algorithm |
|---|---|---|
| Iris | 0.05855103 | 0.05923513 |
| Wine | 0.0080028 | 0.01200976 |

Table 3 shows the result of Jagota index for both k-means and proposed algorithms. The result shows that the Jagota index for the proposed algorithm is less than k-means algorithm.

TABLE 3. The result of Jagota index for k-means and the proposed algorithm

| Data sets | k-means | proposed algorithm |
|---|---|---|
| Iris | 0.7013746 | 0.651203 |
| Wine | 0.597923 | 0.527967 |

As we can see, the result of Dunn and Jagota indexes show that the proposed algorithm is clusters the data better than k-means algorithm because the improved downhill simplex algorithm selects the better fist means. Also One of the most important advantages of the proposed algorithm is seeking the result faster than other algorithms such as k-means. In the following figures we can see that the proposed algorithm is converged faster than k-means. It means that the improved downhill simplex search selects better initial means for clustering the data.

In Fig 7 and Fig 8 the vertical axis indicates the Dunn index and horizontal axis indicates the algorithm iteration for Wins data set.

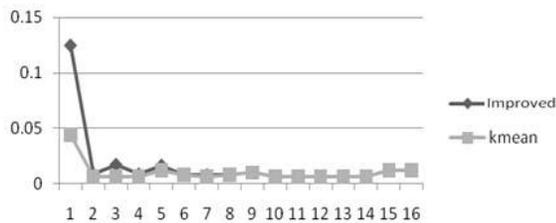

Figure 7. The Dunn index of improved method and k-means for Wine Dataset

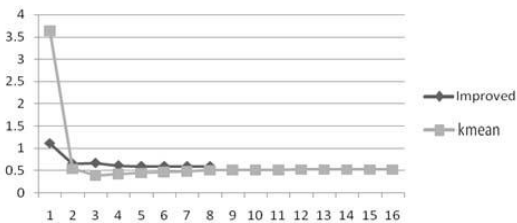

Figure 8. The Jagota index of improved method and k-means for Wine Dataset

In Fig 9 and Fig 10 the vertical axis indicates the Jagota index and horizontal axis indicates the algorithm iteration for Iris data set.

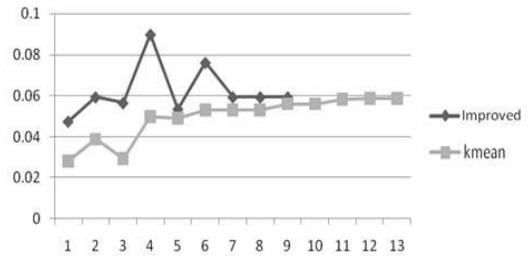

Figure 9. The Dunn index of improved method and k-means for Iris Dataset

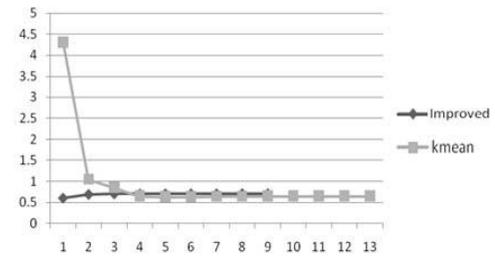

Figure 10. The Jagota index of improved method and k-means for Iris Dataset

## VI. CONCLUSION

In this paper we proposed a new optimization algorithm based on downhill simplex search and compared it with GKM and IGKM algorithms. The results show that the proposed algorithm improves the efficiency of k-means by increasing the accuracy precision of clustering, increasing the Dunn index and decreasing the Jagota index. Also the proposed method is converged so faster than k-means therefore this method is so faster than k-means. The proposed algorithm is a little less accurate than GKM and IGKM but due to its simplicity and speed it has competitive advantages to GKM and IGKM. Therefore the proposed algorithm improved the efficiency and time complexity of k-means algorithm, also this algorithm is so faster than IGKM and GKM with accuracy precision closer to them. So the proposed algorithm can be used as IGKM algorithms in situations that time are too important.